%% file: root.tex
\documentclass[letterpaper, 10 pt, conference]{ieeeconf}  

\IEEEoverridecommandlockouts                              

\usepackage{cite}
\usepackage{graphicx} 
\usepackage{epsfig} 
\usepackage{times} 
\usepackage{amsmath} 
\usepackage{amssymb}  

\usepackage{algorithm}
\usepackage{algorithmicx}
\usepackage[noend]{algpseudocode}
\algnewcommand\algorithmicnot{\textbf{not}}
\algdef{SE}[IF]{IfNot}{EndIf}[1]{\algorithmicif\ \algorithmicnot\ #1\ \algorithmicthen}{\algorithmicend\ \algorithmicif}%
\algtext*{EndIf}
\usepackage{harpoon}
\usepackage{graphicx}
\usepackage{booktabs}  
\usepackage{threeparttable}
\usepackage{multirow}

\usepackage{makecell}
\usepackage{color}
\usepackage{bbding}
\usepackage{stackengine}
\usepackage[hidelinks]{hyperref}
\usepackage{mathtools}

\title{\LARGE \bf
Adaptive Spatio-Temporal Voxels Based Trajectory Planning for Autonomous Driving in Highway Traffic Flow}

\author{Zhiqiang Jian$^{1}$, Songyi Zhang$^{1}$, Lingfeng Sun$^{2}$, Wei Zhan$^{2}$, Masayoshi Tomizuka$^{2}$, and Nanning Zheng$^{1\dag}$ 
\thanks{*This work was supported by the National Natural Science Foundation of China (No.62088102, 61790563).}
\thanks{$^{1}$Z. Jian, S. Zhang, and N. Zheng are with the Institute of Artificial Intelligence and Robotics, Xi'an Jiaotong University, Xi'an, Shaanxi 710049, P.R. China; {\tt\small flztiii, zhangsongyi@stu.xjtu.edu.cn; nnzheng@mail.xjtu.edu.cn}}%
\thanks{$^{2}$L. Sun, W. Zhan, and M. Tomizuka are with the Department of Mechanical Engineering, University of California, Berkeley, CA 94720, USA; {\tt\small lingfengsun, wzhan, tomizuka@berkeley.edu}}%
\thanks{$^{\dag}$N. Zheng is the corresponding author.}%
}

\begin{document}

\maketitle
\thispagestyle{empty}
\pagestyle{empty}

\begin{abstract}

Trajectory planning is crucial for the safe driving of autonomous vehicles in highway traffic flow. Currently, some advanced trajectory planning methods utilize spatio-temporal voxels to construct feasible regions and then convert trajectory planning into optimization problem solving based on the feasible regions. However, these feasible region construction methods cannot adapt to the changes in dynamic environments, making them difficult to apply in complex traffic flow. In this paper, we propose a trajectory planning method based on adaptive spatio-temporal voxels which improves the construction of feasible regions and trajectory optimization while maintaining the quadratic programming form. The method can adjust feasible regions and trajectory planning according to real-time traffic flow and environmental changes, realizing vehicles to drive safely in complex traffic flow. The proposed method has been tested in both open-loop and closed-loop environments, and the test results show that our method outperforms the current planning methods.

\end{abstract}


\input{pages/introduction}

\input{pages/related_works}

\input{pages/method}

\input{pages/experimental_results}

\input{pages/conclusion}

\addtolength{\textheight}{-12cm}   


\bibliographystyle{IEEEtran}
\bibliography{reference}

\end{document}

%% file: pages/introduction.tex
\section{INTRODUCTION}

Highways are suitable application scenarios for autonomous driving \cite{claussmann2019review}. In these scenarios, the key issue is how to deal with the interaction between the ego vehicle and the other agents in the traffic flow. A trajectory planner aims to solve the problem. The trajectory planner can generate a trajectory that meets the kinodynamic constraints of the vehicle in real-time according to the current state of the vehicle and the environment, and ensure the safety, efficiency, and comfort of the vehicle when traveling along the planned trajectory.

Currently, a kind of the state-of-the-art trajectory planning methods searches for a feasible region in the spatio-temporal domain, and all the trajectories belonging to it can guarantee the safety of the vehicle. Afterward, an optimization problem is defined according to the feasible region, and its solution uniquely corresponds to the planned trajectory. For example, Ding \emph{et al.} \cite{ding2019safe} propose the Spatio-Temporal Semantic Corridor (SSC) to build a feasible region based on the results of behavior planning, and then define an optimization problem and solve it to plan a trajectory. Similarly, Zhang \emph{et al.} \cite{zhang2021unified} propose the Spatio-Temporal Voxels (STV) for feasible region construction and also define an optimization problem under the constraints of the feasible region to obtain a trajectory. These methods have achieved good performance.

However, these methods still have some shortcomings. Ding \emph{et al.}'s method relies on the behavior planner, and unqualified behavior planning results will lead to unqualified planned trajectories. Zhang \emph{et al.}'s method does not have the above problem, but its feasible region construction and trajectory optimization are not flexible enough. Zhang \emph{et al.}'s method use voxels to describe the feasible region, but the number and size of the voxels are fixed and the lateral kinodynamic constraints of the vehicle are ignored. Besides, its objective function only considers jerks, which is insufficient. These problems make the method perform poorly in more dense traffic flows. 

\begin{figure}[!t]
    \centering
    \includegraphics[width=\linewidth]{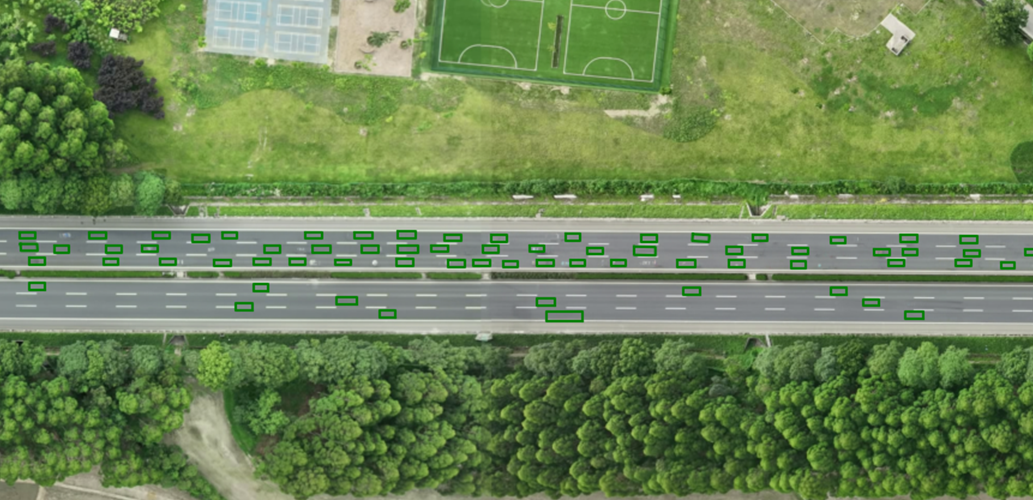}
    \caption{This figure shows one frame of the traffic flow from the CitySim dataset. The green boxes indicate the annotated vehicles.}
    \label{fig:freeway}
\end{figure}

In order to solve the problems, this paper proposes an Adaptive Spatio-Temporal Voxels Based Trajectory Planning (ASVP) method. The ASVP method improves the method proposed by Zhang \emph{et al.}: In terms of the feasible region construction, the ASVP method uses multi-resolution voxels to construct the feasible region, and the number of the voxels in the feasible region can also change according to the complexity of the environment. In terms of the optimization problem definition, the objective function and the constraints are both modified, and more factors are considered in the optimization. Based on the improvements, the proposed method can ensure the safe, efficient, and comfortable traveling of the vehicle in complex traffic flows.

\begin{figure*}[!t]
    \centering
    \includegraphics[width=0.96\linewidth]{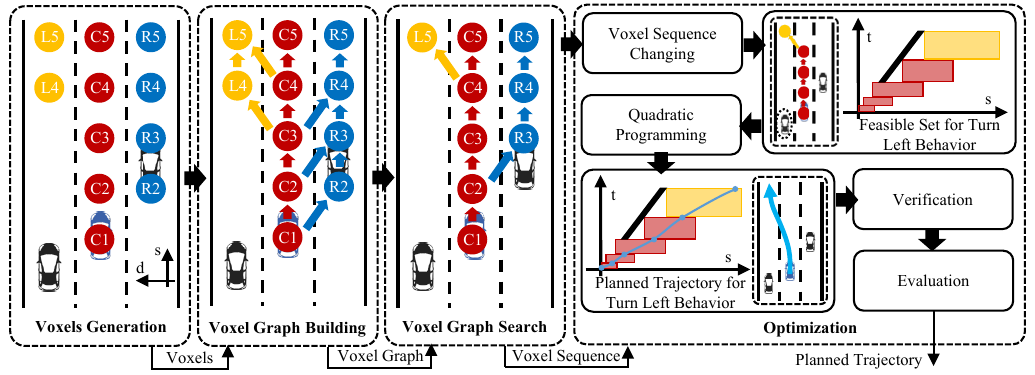}
    \caption{This figure shows the framework of the proposed method, which includes four modules. In the first three modules, the blue vehicle is the ego vehicle. The colored circles indicate the voxels and different colors represent different lanes. The colored arrows indicate the edges of the voxel graph. In the last module, the colored boxes are the voxels in the spatio-temporal domain. The black region indicates the marked obstacle vehicle in the spatio-temporal domain. The blue curve indicates the planned trajectory.}
    \label{fig:framework}
\end{figure*}

The proposed method is tested in both open-loop and closed-loop environments. The open-loop tests are conducted by replaying the CitySim dataset \cite{https://doi.org/10.48550/arxiv.2208.11036}, as shown in Fig.~\ref{fig:freeway}. The closed-loop environment uses the gym simulator \cite{highway-env}. In both tests, the proposed method achieves significant performance in the safety and efficiency of the vehicle. Besides, an ablation study is designed, which verifies the effectiveness of the innovations proposed in the paper.

In conclusion, the contributions of the paper are as follows.
\begin{itemize}
    \item A novel feasible region construction method is proposed, which can be adjusted according to the environment and adapted to scenarios of different complexity.
    \item The defined optimization problem considers the safety, efficiency, and comfort of the vehicle while maintaining the quadratic programming form.
    \item Combining the above innovations, an effective highway traffic flow trajectory planning method is implemented, which is open source. \footnote{\url{https://github.com/flztiii/traffic_flow_trajectory_planning}}
\end{itemize}

%% file: pages/related_works.tex
\section{RELATED WORKS}

One of the main methods of trajectory planning is to convert the planning problem into an optimization problem \cite{gonzalez2015review}. Some methods try to directly solve the optimization problem, such as the Sequential Quadratic Programming (SQP) method applied by Ziegler \emph{et al.} \cite{ziegler2014trajectory}. However, since the optimization problem is non-convex and has high complexity, more methods simplify the problem before solving it, including sampling and evaluation methods \cite{werling2010optimal, chu2012local, li2017development, zhang2021trajectory}, spatial and temporal domain separation methods \cite{xu2012real, villagra2012smooth, fan2018baidu, lim2019hybrid}, and direct spatio-temporal domain simplification methods \cite{ziegler2009spatiotemporal, ding2019safe, ding2021epsilon, zhang2021unified}.

The sampling and evaluation methods sample some trajectories, and then evaluate these trajectories through the objective function. The trajectory with the lowest cost is selected as the planning trajectory. For example, Werling \emph{et al.} \cite{werling2010optimal} use polynomial curves to generate sampling trajectories in the Frenet coordinate system, and then evaluate them through the objective function to achieve trajectory planning. Zhang \emph{et al.} \cite{zhang2021trajectory} also apply polynomial curves to generate paths and velocity profiles to obtain sampling trajectories, and finally complete trajectory planning through evaluation.

The spatial and temporal domain separation methods plan separately in the spatial and temporal domains, and then combine the two to obtain the planned trajectory. For example, Xu \emph{et al.} \cite{xu2012real} use the sampling-evaluation method to obtain the separate initial solutions from the spatial and temporal domains, and then use the numerical optimization method to obtain the planned trajectory. Fan \emph{et al.} \cite{fan2018baidu} propose a method similar to Expectation-Maximization (EM) to iteratively optimize the trajectory in the separated spatial and temporal domains, thereby obtaining the planned trajectory.

The direct spatio-temporal domain simplification methods construct a simplified feasible region in the spatio-temporal domain according to the environment and the vehicle's kinodynamic constraints and then solve the optimization problem in the feasible region. For example, Ziegler et al. \cite{ziegler2009spatiotemporal} apply the state lattice method to discretize the spatio-temporal domain, thus realizing the simplification. The SSC proposed by Ding et al. \cite{ding2019safe} and the STV proposed by Zhang et al. \cite{zhang2021unified} also obtain a simplified feasible region, so as to optimize the trajectory accordingly. Currently, this kind of method achieves the best performance in highway trajectory planning. The method proposed in this paper is also carried out following the idea.

%% file: pages/method.tex
\section{METHOD}

\subsection{Framework}

The proposed method's framework is shown in Fig.~\ref{fig:framework}. The method's framework is inspired by Zhang \emph{et al.}'s method \cite{zhang2021unified}, but we have modified the framework's modules. The planner updates at 5 $\mathrm{Hz}$ frequency and outputs the trajectory in the future time period $T$ to the control module in real-time. In each planning episode, the planner conducts feasible region construction and trajectory optimization. These two processes are carried out in the Frenet coordinate system and the ego vehicle and other agent information are all converted to the Frenet coordinate system \cite{ding2021epsilon}. In this case, the spatio-temporal domain can be described by the $s$-$d$-$t$ coordinate system. The $s$ axis is the lane's direction, and the $d$ axis is perpendicular to the lane's direction.

In the feasible region construction, spatio-temporal voxels are generated according to the environment. Each voxel corresponds to a lane $l$, where $l \in \{\mathrm{L, C, R}\}$. Lane $\mathrm{L}$ indicates the left lane, $\mathrm{R}$ indicates the right lane, and $\mathrm{C}$ indicates the current lane. Afterward, a voxel graph is built and each voxel's cost is calculated. For each behavior, a voxel sequence with the least cost is searched in the voxel graph. Behaviors include left lane change, right lane change, and lane keeping.

In the trajectory optimization, each voxel sequence is modified according to its corresponding behavior, and a feasible region can be obtained accordingly. According to the feasible region, a quadratic programming problem can be defined to describe the trajectory planning problem. The quadratic programming problem's solution uniquely corresponds to the planned trajectory. Then, the planned trajectory is verified. If the verification fails, the voxel sequence will be changed and the quadratic programming will be performed again. If the verification succeeds, the planned trajectory's cost is evaluated. Finally, for each behavior, a trajectory with the cost is planned. The trajectory with the lowest cost is output to the control module.

\subsection{Feasible Region Construction}

We discretize the future time period $T$ into $n$ segments $\{\Delta T_i\}_{i=0}^{n-1}$. for $\forall i$, $\Delta T_{i+1} \geq \Delta T_i$. For each lane $l$ in each time segment $\Delta T_i$, a voxel set $\{\mathbf{v}_{l,i,j}\}_{j=0}^{m-1}$ will be generated. Each voxel $\mathbf{v}$ represents a box region in the spatio-temporal domain and is defined as $\mathbf{v}=(ls, us, ld, ud, lt, ut)^\mathrm{T}$. $us$ and $ls$ are the voxel $\mathbf{v}$'s upper and lower bounds on the $s$ axis. $ud$ and $ld$ are the voxel $\mathbf{v}$'s upper and lower bounds on the $d$ axis. $ut$ and $lt$ are the voxel $\mathbf{v}$'s upper and lower bounds on the $t$ axis. For $\forall i$, there exists $ut_i - lt_i=\Delta T_i$ and $ut_i=lt_{i+1}$.

\textbf{Voxels generation.} The generated voxel set is defined as $\{\mathbf{v}_{l,i,j}\}_{j=0}^{m-1}$. For $\forall j$, the voxel $\mathbf{v}_{l,i,j}$'s the upper and lower bounds on the $t$ axis remain unchanged and are denote as $ut_{l,i}$ and $lt_{l,i}$, which can be calculated according to the time segments $\{\Delta T_i\}_{i=0}^{n-1}$.

Then, the voxel $\mathbf{v}_{l,i,j}$'s upper and lower bounds $us_{l,i,j}$ and $ls_{l,i,j}$ on the $s$ axis are calculated. According to the vehicle's initial longitudinal position $s_0$ and velocity $v_0^{(s)}$, the reachable lower bound $s_{l,i}^\mathrm{(min)}$ of the $s$ axis when the vehicle decelerates at the maximum deceleration within the time $lt_{l,i}$ and the reachable upper bound $s_{l,i}^\mathrm{(max)}$ of the $s$ axis when the vehicle accelerates at the maximum acceleration within the time $ut_{l,i}$ can be obtained \cite{zhang2021unified}. For each lane $l$, its related agent set is defined as $\mathcal{O}_l$. (For example, if $l=\mathrm{L}$, $\mathcal{O}_l$ will be the agents on the left lane.) For $\forall o \in \mathcal{O}_l$, its occupied range of the $s$ axis within time $lt_{l,i}$ to $ut_{l,i}$ can be obtained, which is defined as $[s_o^\mathrm{(min)}, s_o^\mathrm{(max) }]$. In this case, the free range set $\mathcal{S}_{l,i}$ can be calculated as follows.
\vspace*{-0.2\baselineskip}
\begin{equation}
    \label{eq:s_range}
    \mathcal{S}_{l,i} = [s_{l,i}^\mathrm{(min)}, s_{l,i}^\mathrm{(max)}] \setminus \bigcup_{o \in \mathcal{O}_l} [s_o^\mathrm{(min)}, s_o^\mathrm{(max)}].
\end{equation}
\vspace*{-0.2\baselineskip}
$\mathcal{S}_{l,i}$ contains multiple free ranges of the $s$ axis. Each free range corresponds to a voxel and the free range's number in $\mathcal{S}_{l,i}$ is $m$. The $j_\mathrm{th}$ free range's upper and lower bounds are $us_{l,i,j}$ and $ls_{l,i,j}$.

At last, the voxel $\mathbf{v}_{l,i,j}$'s upper and lower bounds $ud_{l,i,j}$ and $ld_{l,i,j}$ on the $d$ axis are calculated. The $\mathbf{v}_{l,i,j}$'s upper and lower bounds on the $d$ axis are related to its lane $l$. If $l=\mathrm{L}$ or $l=\mathrm{R}$, $ud_{l,i,j}$ and $ld_{l,i,j}$ will be the corresponding lane boundary's position shrunk by the vehicle's width. If $l=\mathrm{C}$, according to the vehicle's initial lateral position $d_0$ and velocity $v_0^{(d)}$, its reachable lateral range will be calculated, which determines $ud_{l,i,j}$ and $ld_{l,i ,j}$. Through the above process, all the voxels in the time period $T$ can be generated, defined as $\mathcal{V}=\{\mathbf{v}_{l,i,j} | l \in \{\mathrm{L},\mathrm{C},\mathrm{R}\}, i \in [0, n-1], j \in [0, m-1]\}$.

\begin{algorithm}[!t]  
    \caption{Voxel Graph Building}  
    \label{alg:voxel_graph}  
    \begin{algorithmic}[1]
    \Require  
    Voxel set $\mathcal{V}$
    \Ensure
    Voxel graph $\mathcal{G}$
    \For{$i=0,\cdots,n-1$}
        \State $\mathcal{G}_i$ $\leftarrow$ $\emptyset$
        \For{$\mathbf{v}$ $\in$ $\{\mathbf{v}_{l,i,j} | l \in \{\mathrm{L},\mathrm{C},\mathrm{R}\}, j \in [0, m-1]\}$}
            \State Initialize a node $\eta$ with the voxel $\mathbf{v}$
            \IfNot{$i = 0$}
                \For{$\eta_\mathrm{f}$ $\in$ $\mathcal{G}_{i-1}$}
                    \IfNot{LaneCheck($\eta$, $\eta_\mathrm{f}$)}
                        \State \textbf{continue} 
                    \EndIf
                    \IfNot{IntersectionCheck($\eta$, $\eta_\mathrm{f}$)}
                        \State \textbf{continue}
                    \EndIf
                    \State Push $\eta_\mathrm{f}$ into $\eta$'s father set $\mathcal{G}_\mathrm{f}$
                    \State Push CalcCost($\eta$, $\eta_\mathrm{f}$) into $\eta$'s cost set $\mathcal{C}$
                \EndFor
            \EndIf
            \State Push $\eta$ into $\mathcal{G}_i$
        \EndFor
        \State Push $\mathcal{G}_i$ into $\mathcal{G}$
    \EndFor
    \end{algorithmic}
    
\end{algorithm}
\textbf{Voxel graph building}. The voxel graph $\mathcal{G}$ contains $n$ layers and each layer $\mathcal{G}_i$ contains multiple nodes $\eta$. The node $\eta$ consists of the voxel $\mathbf{v}$, the cost set $\mathcal{C}$ and the parent node set $\mathcal{G}_\mathrm{f}$. The voxel graph's building is shown in Alg.~\ref{alg:voxel_graph}. For the nodes in the first layer $\mathcal{G}_0$, since there is no need to consider its cost set and parent node set, the nodes can be directly defined according to the voxels.

For the nodes in the $i_\mathrm{th}$ layers, their definition is as follows. First, for each voxel $\mathbf{v}$ in the voxel set $\{\mathbf{v}_{l,i,j} | l \in \{\mathrm{L},\mathrm{C},\mathrm{R}\}, j \in [0, m-1]\}$, a node $\eta$ is defined accordingly. Each node $\eta_\mathrm{f}$ in the previous layer $\mathcal{G}_{i-1}$ is judged whether it is the parent node of $\eta$. Judgment conditions include lane checking and intersection checking. In the lane checking, it is judged whether the lane of the node $\eta$'s voxel is the current lane. If so, the lane of the node $\eta_\mathrm{f}$'s voxel must also be the current lane. If not, the lane of the node $\eta_\mathrm{f}$'s voxel must be the same as $\eta$, or be the current lane. The lane checking aims to prevent multiple lane changes during one planning episode. In the intersection checking, it is judged whether the overlapping of the nodes $\eta$ and $\eta_\mathrm{f}$'s voxels on the $s$ axis and $d$ axis is larger than the threshold. If the lane checking and intersection checking are satisfied, the node $\eta_\mathrm{f}$ will be added into the node $\eta$'s parent node set $\mathcal{G}_\mathrm{f}$. Finally, the cost $c$ is calculated according to the nodes $\eta$ and $\eta_\mathrm{f}$'s voxels and added to the node $\eta$'s cost set $\mathcal{C}$. The calculation is as follows.
\vspace*{-0.2\baselineskip}
\begin{equation}
    \label{eq:node_cost}
    c = 1 - \frac{2 s_\mathrm{inter}}{\Delta T_i^2 (a_\mathrm{max}^{(s)} - a_\mathrm{min}^{(s)})},
\end{equation}
\vspace*{-0.2\baselineskip}
where $s_\mathrm{inter}$ is the overlapping length of the nodes $\eta$ and $\eta_\mathrm{f}$'s voxels on the $s$ axis, $a_\mathrm{max}^{(s)}$ is the vehicle's longitudinal maximum acceleration, $a_\mathrm{min}^{(s)}$ is the vehicle's longitudinal maximum deceleration. The cost reflects how much other agents limit the vehicle's motion. When $c=0$, other agents have no restrictions on the ego vehicle's motion, and when $c=1$, other agents completely restrict the ego vehicle's motion.

\textbf{Voxel graph search.} After building the voxel graph $\mathcal{G}$, the final step is to generate the voxel sequence $\mathcal{V}_b=\{\mathbf{v}_{l,i}\}_{i=0}^{n-1}$ for each behavior $b$. Nodes that correspond to the behavior $b$ are selected in the voxel map $\mathcal{G}$'s last layer $\mathcal{G}_{n-1}$. For example, if the behavior $b$ is left lane change, the nodes whose voxels' corresponding lane is the left lane will be selected. For each selected node, the depth-first search is applied to get the paths connecting it and the nodes in the first layer $\mathcal{G}_0$. The path with the least cost will be chosen as $\mathcal{V}_{b}$. Through the above process, the voxel sequences corresponding to all three behaviors can be obtained.

\subsection{Trajectory Optimization}

\textbf{Voxel sequence change.} For each behavior $b$, a trajectory $\Xi_{b}$ is planned according to the voxel sequence $\mathcal{V}_{b}=\{\mathbf{v}_{l,i}\}_{i=0}^{n-1}$. If the behavior $b$ is not lane keeping, the voxel sequence $\mathcal{V}_{b}$ needs to be changed and the process is as follows.

The index $i$ in the voxel sequence $\mathcal{V}_{b}$ is found, so that voxel $\mathbf{v}_{l,i}$ and $\mathbf{v}_{l ,i+1}$ have different lanes. In the voxel graph $\mathcal{G}$'s $i+1_\mathrm{th}$ layer $\mathcal{G}_{i+1}$, the voxels that hold the same lane as the voxel $\mathbf{v}_{l,i}$ are selected. These voxels are intersected with the voxel $\mathbf{v}_{l,i}$ on the $s$ axis separately, and the maximum range is regarded as the voxel $\mathbf{v}_{l,i}$'s new upper and lower bounds on the $s$ axis. Similarly, in the voxel graph $\mathcal{G}$'s $i_\mathrm{th}$ layer $\mathcal{G}_{i}$, the voxels that hold the same lane as the voxel $\mathbf{v}_{l,i+1}$ are selected. These voxels are intersected with the voxel $\mathbf{v}_{l,i+1}$ on the $s$ axis separately, and the maximum range is regarded as the voxel $\mathbf{v}_{l,i+1}$'s new upper and lower bounds on the $s$ axis. Afterward, The voxel $\mathbf{v}_{l,i}$ and $\mathbf{v}_{l,i+1}$’s union on the $d$ axis is regarded as the two voxels' new upper and lower bounds on the $d$ axis. The voxel sequence change aims to allow the planner to consider the agents on both lanes at the same time in the process of changing lanes to ensure safety.

\textbf{Optimization definition.} an optimization problem is defined based on the voxel sequence $\mathcal{V}_{b}$ to achieve trajectory planning. The key to optimization problem definition includes optimization variables, constraints, and an objective function. For the optimization variables, the planned trajectory is described by a piece-wise quintic Bezier curve $\Xi(t)=\{\xi_i^{(\sigma)}(t) | i \in [0,n-1], \sigma \in \{s, d\}\}$, whose expression is as follows.
\vspace*{-0.2\baselineskip}
\begin{equation}
    \centering
    \label{eq:bezier}
    \begin{aligned}
        &\xi_i^{(\sigma)}(t) = \Delta T_i B_i^{(\sigma)}(\frac{t - lt_{l,i}}{\Delta T_i}), \quad t \in [lt_{l,i}, ut_{l,i}], \\
        &B_i^{(\sigma)}(x) = \sum_{k=0}^5 \begin{pmatrix} 5 \\ k \end{pmatrix} p_{i,k}^{(\sigma)} (1-x)^{(5-k)} x^k, \\
    \end{aligned}
\end{equation}
\vspace*{-0.2\baselineskip}
where $lt_{l,i}$ and $ut_{l,i}$ are the voxel $\mathbf{v}_{l,i}$'s upper and lower bounds on the $t$ axis and $\Delta T_i = ut_{l,i} - lt_{l,i}$. $\mathcal{P}=\{p_{i,k}^{(\sigma)} | \sigma \in \{s, d\}, k \in [0, 5], i \in [0, n-1]\}$ are the piece-wise Bezier curve's control points. As long as $\mathcal{P}$ is determined, the piece-wise Bezier curve $\Xi(t)$ can be uniquely determined. Therefore, $\mathcal{P}$ can be regarded as the optimization variables.

The constraints include equality constraints and inequality constraints. Among the equality constraints, the first consideration is that the planned trajectory's initial configuration needs to be the same as the ego vehicle's current configuration, so the following equation needs to be satisfied.
\vspace*{-0.2\baselineskip}
\begin{equation}
    \label{eq:init_constraints}
    \begin{aligned}
        &\xi_0^{(s)}(lt_{l,0}) = s_0, \quad \xi_0^{(d)}(lt_{l,0}) = d_0, \\
        \frac{\partial \xi_0^{(\sigma)}(lt_{l,0})}{\partial t} &= v_0^{(\sigma)}, \frac{\partial^2 \xi_0^{(\sigma)}(lt_{l,0})}{\partial t^2} = a_0^{(\sigma)}, \sigma \in \{s, d\}, \\
    \end{aligned}
\end{equation}
\vspace*{-0.2\baselineskip}
where $s_0$ and $d_0$ are the vehicle's initial longitudinal and lateral positions, $v_0^{(s)}$ and $v_0^{(d)}$ are the vehicle's initial longitudinal and lateral velocities, and $a_0^{(s)}$ and $a_0^{(d)}$ are the vehicle's initial longitudinal and lateral accelerations. Among the equality constraints, the second consideration is that the planned trajectory needs to meet the $\mathrm{C}^2$ continuity, so the following equation needs to be satisfied.
\vspace*{-0.2\baselineskip}
\begin{equation}
    \label{eq:g2_constraints}
    \begin{aligned}
        &\frac{\partial^\mu \xi_i^{(\sigma)}(ut_{l,i})}{\partial t^\mu}  = \frac{\partial^\mu \xi_{i+1}^{(\sigma)}(lt_{l,i+1})}{\partial t^\mu}, \\
        \mu=&0,1,2, \quad \sigma \in \{s, d\}, \quad i \in [0, n-2], \\
    \end{aligned}
\end{equation}
\vspace*{-0.2\baselineskip}
where $\frac{\partial^\mu \xi_i^{(\sigma)}(t)}{\partial t^\mu}$ is the function $\xi_i^{(\sigma)}(t)$'s $\mu$ order derivative.

For the inequality constraints, safety and comfort are considered. The constraint brought by safety is that the Bezier curve $\Xi(t)$ must be contained in the voxel sequence. The constraints brought by comfort are that the Bezier curve $\Xi(t)$'s first-order, second-order, and third-order derivatives, which represent the limits on velocity, acceleration, and jerk respectively, must be within the limit range. Due to the Bezier curve's convex hull property, as long as the Bezier curve's control points are within the limit range, all the points on the Bezier curves can be guaranteed to be within the limit range \cite{ding2019safe}. Therefore, the safety inequality constraints are as follows.
\vspace*{-0.2\baselineskip}
\begin{equation}
    \label{eq:safe_neq_constraints}
    \begin{aligned}
        ls_{l,i} \leq p^{(s)}_{i,k} \leq &us_{l,i}, \quad ld_{l,i} \leq p^{(d)}_{i,k} \leq ud_{l,i}, \\
        &i \in [0, n-1], \\
    \end{aligned}
\end{equation}
\vspace*{-0.2\baselineskip}
where $ls_{l,i}$ and $us_{l,i}$ are the voxel $\mathrm{v}_{l,i}$'s upper and lower bounds on the $s$ axis, and $ld_{l,i}$ and $ud_{l,i}$ are the voxel $\mathrm{v}_{l,i}$'s upper and lower bounds on the $d$ axis. The comfort inequality constraints are as follows.
\vspace*{-0.2\baselineskip}
\begin{equation}
    \label{eq:comfort_neq_constraints}
    \begin{aligned}
        &v_\mathrm{min}^{(\sigma)} \leq p^{(\sigma)}_{i,k}{}' \leq v_\mathrm{max}^{(\sigma)}, \quad k \in [0, 4], \\
        &a_\mathrm{min}^{(\sigma)} \leq p^{(\sigma)}_{i,k}{}'' \leq a_\mathrm{max}^{(\sigma)}, \quad k \in [0, 3], \\
        &j_\mathrm{min}^{(\sigma)} \leq p^{(\sigma)}_{i,k}{}''' \leq j_\mathrm{max}^{(\sigma)}, \quad k \in [0, 2], \\
        &i \in [0, n-1], \quad \sigma \in \{s,d\}, \\
    \end{aligned}
\end{equation}
\vspace*{-0.2\baselineskip}
where $v_\mathrm{min}^{(s)}$ and $v_\mathrm{max}^{(s)}$ are the vehicle's longitudinal velocity upper and lower bounds, $v_\mathrm{min}^{(d)}$ and $v_\mathrm{max}^{(d)}$ are the vehicle's lateral velocity upper and lower bounds, $a_\mathrm{min}^{(s)}$ and $a_\mathrm{max}^{(s)}$ are the vehicle's longitudinal acceleration upper and lower bounds, $a_\mathrm{min}^{(d)}$ and $a_\mathrm{max}^{(d)}$ are the vehicle's lateral acceleration upper and lower bounds, $j_\mathrm{min}^{(s)}$ and $j_\mathrm{max}^{(s)}$ are the vehicle's longitudinal jerk upper and lower bounds, and $j_\mathrm{min}^{(d)}$ and $j_\mathrm{max}^{(d)}$ are the vehicle's lateral jerk upper and lower bounds. $p^{(\sigma)}_{i,k}{}'$, $p^{(\sigma)}_{i,k}{}''$ and $p^{(\sigma)}_{i,k}{}'''$ are the Bezier curve $\Xi(t)$'s first-order, second order, and third-order derivatives' control points (A Bezier curve's derivative is still a Bezier curve), which are the linear combination of the Bezier curve $\Xi(t)$'s control points \cite{farin2014curves}.

The objective function is defined as $\sum_{j=0}^4 w_j \varepsilon_j$, which is weighted sum of five terms. The term $\varepsilon_0$ is calculated as follows.
\vspace*{-0.2\baselineskip}
\begin{equation}
    \label{eq:e0}
    \varepsilon_0 = \sum_{\sigma \in \{s,d\}} \sum_{i=0}^{n-1} \int_{lt_{l,i}}^{ut_{l,i}} \Big{(}\frac{\partial^3 \xi_i^{(\sigma)}(t)}{\partial t^3}\Big{)}^2 \mathrm{d}t.
\end{equation}
\vspace*{-0.2\baselineskip}
The term $\varepsilon_0$'s purpose is to minimize the jerk. The term $\varepsilon_1$ is calculated as follows.
\vspace*{-0.2\baselineskip}
\begin{equation}
    \label{eq:e1}
    \varepsilon_1 = \sum_{\sigma \in \{s,d\}} \sum_{i=0}^{n-1} \Big{(}\xi_i^{(\sigma)}(ut_{l,i}) - \alpha_i^{(\sigma)}\Big{)}^2,
\end{equation}
\vspace*{-0.2\baselineskip}
where $\alpha_i^{(s)}$ and $\alpha_i^{(d)}$ are the ideal longitudinal and lateral end positions of the Bezier curve's $i_\mathrm{th}$ segment. The term $\varepsilon_1$ aims to make the end position of the Bezier curve's each segment close to the expectation. $\alpha_i^{(s)}$ and $\alpha_i^{(d)}$'s calculation are as follows. The front and rear agents of the bounding voxel $\mathbf{v}_{l,i}$ on the lane $l$ are found from time $lt_{l,i}$ to $ut_{l,i}$. If the front agent does not exist, $\alpha_i^{(s)}$ will be set to the voxel $\mathbf{v}_{l,i}$'s upper bound on the $s$ axis. If the front and rear agents exist, the front agent's position $s_\mathrm{f}$ on the $s$ axis at time $lt_{l,i}$ and the rear agent's position $s_\mathrm{r}$ on the $s$ axis at time $ut_{l,i}$ will be obtained. Then, $\alpha_i^{(s)}$ can be calculated as follows.
\vspace*{-0.2\baselineskip}
\begin{equation}
    \label{eq:expected_pos}
    \begin{aligned}
        &\gamma_\mathrm{f} = s_\mathrm{f} + \frac{(v^{(s)}_0)^2 - (v^{(s)}_\mathrm{f})^2}{2a^{(s)}_\mathrm{max}} - v^{(s)}_0 T_\mathrm{res}, \\ 
        &\gamma_\mathrm{r} = s_\mathrm{r} + \frac{(v^{(s)}_\mathrm{r})^2 - (v^{(s)}_0)^2}{2a^{(s)}_\mathrm{max}} + v^{(s)}_\mathrm{r} (T_\mathrm{res} + ut_{l,i} - lt_{l,i}), \\ 
        &\alpha_i^{(s)} = \left\{
        \begin{array}{l}
            \gamma_\mathrm{f}, \quad \gamma_\mathrm{r} \leq \gamma_\mathrm{f}, \\
            \frac{w_\mathrm{f}}{w_\mathrm{f} + w_\mathrm{r}} \gamma_\mathrm{f} + \frac{w_\mathrm{r}}{w_\mathrm{f} + w_\mathrm{r}} \gamma_\mathrm{r}, \quad \mathrm{else}, \\
        \end{array}  
        \right.\\
    \end{aligned}
\end{equation}
\vspace*{-0.2\baselineskip}
where $v^{(s)}_0$ is the ego vehicle's initial longitudinal velocity, $v^{(s)}_\mathrm{f}$ is the front agent's longitudinal velocity, $v^{(s)}_\mathrm{r}$ is the rear agent's longitudinal velocity, $T_\mathrm{res}$ is the preset response time, $a^{(s)}_\mathrm{max}$ is the vehicle's maximum longitudinal deceleration, and $w_\mathrm{f}$ and $w_\mathrm{r}$ are preset weights. If the rear agent does not exist, $\alpha_i^{(s)}$ equals to $\gamma_\mathrm{f}$. $\alpha_i^{(d)}$ is determined by the voxel $\mathbf{v}_{l,i}$'s lane. For example, if the voxel $\mathbf{v}_{l,i}$'s lane is the left lane, $\alpha_i^{(d)}$ equals to the lane width. The term $\varepsilon_2$ is calculated as follows.
\vspace*{-0.2\baselineskip}
\begin{equation}
    \label{eq:e2}
    \varepsilon_2 = \sum_{\sigma \in \{s,d\}} \sum_{i=0}^{n-1} \Big{(}\frac{\partial \xi_i^{(\sigma)}(ut_{l,i})}{\partial t} - \beta_i^{(\sigma)}\Big{)}^2,
\end{equation}
\vspace*{-0.2\baselineskip}
where $\beta_i^{(s)}$ and $\beta_i^{(d)}$ are the ideal end velocities of the Bezier curve's $i_\mathrm{th}$ segment. $\beta_i^{(s)}$ is also determined by the voxel $\mathbf{v}_{l,i}$'s front agent on the lane $l$ from time $lt_{l,i}$ to $ut_{l,i}$. If the front agent does not exist, $\beta_i^{(s)}$ will be set as the upper limit of velocity allowed by kinematics. If the front agent exists, $\beta_i^{(s)}$ will be set to the front agent's longitudinal velocity. $\beta_i^{(d)}$ is set to zero. The term $\varepsilon_3$ is calculated as follows.
\vspace*{-0.2\baselineskip}
\begin{equation}
    \label{eq:e3}
    \varepsilon_3 = \sum_{i=0}^{n-1} \int_{lt_{l,i}}^{ut_{l,i}} \Big{(}\frac{\partial \xi_i^{(d)}(t)}{\partial t}\Big{)}^2 \mathrm{d}t.
\end{equation}
\vspace*{-0.2\baselineskip}
The purpose of the term $\varepsilon_3$ is to make the lateral velocity close to constant and the lateral position change uniformly. The term $\varepsilon_4$ is calculated as follows.
\vspace*{-0.2\baselineskip}
\begin{equation}
    \label{eq:e4}
    \varepsilon_4 = \sum_{i=0}^{n-1} \int_{lt_{l,i}}^{ut_{l,i}} \Big{(}\frac{\partial^2 \xi_i^{(s)}(t)}{\partial^2 t}\Big{)}^2 \mathrm{d}t.
\end{equation}
\vspace*{-0.2\baselineskip}
The term $\varepsilon_4$ aims to make the longitudinal acceleration close to constant and the longitudinal velocity change uniformly.

\textbf{Optimization solver.} It can be easily seen through the calculation that the defined constraints are linear, the defined objective function is quadratic, and its quadratic coefficient matrix is semi-positive. These properties show that the defined optimization problem is quadratic programming. Therefore, Object-Oriented Software for Quadratic Programming (OOQP) \cite{gertz2003object} can be applied to solve the problem to obtain the planned trajectory. The planned trajectory is verified whether it is acceptable. The verification conditions are whether the trajectory meets all the constraints and whether the trajectory's curvatures are smaller than the threshold. 

\textbf{Verification and evaluation.} If the optimization fails or the planned trajectory fails to pass the verification, the voxel sequence $\mathcal{V}_b$ will be changed again. We continuously remove the tail voxel of the voxel sequence and repeat the optimization until the planned trajectory passes the verification. Due to the existence of $\Delta T_{i+1} \geq \Delta T_i$, the tail voxel of the voxel sequence has stronger constraints in the optimization. Therefore, as the tail voxel of the voxel sequence is kept removed, the optimization problem becomes increasingly easier to be solved. As the above process continues, the planned trajectory will become more and more aggressive. After the trajectories for each behavior are planned, the trajectories are evaluated. The evaluation method uses Eq.~\ref{eq:node_cost} to calculate the cost of the voxel sequence corresponding to the trajectory, and the trajectory with the lowest cost is output to the controller.

%% file: pages/experimental_results.tex
\section{EXPERIMENTAL RESULTS}

The experiment includes the open-loop tests, closed-loop tests, and ablation study. The purpose of the open-loop tests is to compare the proposed method, the SOTA method, and human drivers through a large number of tests under safety and efficiency and proves the proposed method's effectiveness. The purpose of the closed-loop experiment is to conduct a long-term test to demonstrate the proposed method's stability. In the process, we analyzed the planning process using the proposed method, and further verified the effectiveness of the method. The ablation study aims to verify whether the proposed innovations can improve the planner. In the experiment, the proposed method is implemented based on C++ and Robot Operating System (ROS) \cite{quigley2009ros}, and runs on a computer equipped with a CPU i7-8750H. The vehicle's maximum longitudinal acceleration and deceleration are 2 $\mathrm{m/s}^2$, maximum lateral acceleration and deceleration are 2 $\mathrm{m/s}^2$, and maximum longitudinal and lateral jerks are 2 $\mathrm{m/s}^3$. The parameters used by the method are detailed in the open source code.

\subsection{Open-Loop Tests}

In the open-loop tests, the CitySim dataset is applied \cite{https://doi.org/10.48550/arxiv.2208.11036}. Compared with the NGSIM dataset, the CitySim dataset has higher vehicle density and complexity.

For the experiment design, Sun \emph{et al.}'s method \cite{sun2021move} is referred to. 10 seconds of traffic flow is randomly selected from the CitySim dataset, from which an agent is randomly selected as the ego vehicle. The agent's state is regarded as the ego vehicle's initial state, and the lane the agent is in after 10 seconds is regarded as the target lane. Afterward, the agent is removed from the traffic flow and a trajectory planning method is used to update the ego vehicle's state in the traffic flow. The states of other agents in the traffic flow are updated according to the CitySim dataset. The update frequency is 5 Hz and the update duration is 10 seconds. During the planning process, the planning method only considers other agents within 100 meters of its own lane and adjacent lanes. For the case where the target lane is the same as the current lane (lane keeping) and the case where the target lane is different from the current lane (lane change), each one is carried out 100 times respectively, and the \emph{success rate}, \emph{failure rate}, \emph{risk}, and \emph{efficiency} are recorded.

The \emph{success rate} is defined as the ratio of the number of times the ego vehicle does not collide with other agents and is in the target lane at the end of the test to the total number of tests. The \emph{failure rate} is defined as the ratio of the number of times the ego vehicle collided with other agents or failed to plan to the total number of tests. The \emph{risk} is defined as the ratio of the time the vehicle is in danger to the total testing time. The danger is defined as the situation in which the ego vehicle's available response time is less than 1 second. The response time is defined that when the front agent brakes with the maximum deceleration, the ego vehicle can brake with the maximum deceleration after the response time, so as to ensure that the ego vehicle will not collide with the front agent. In common sense, a response time of 1 second is enough to ensure the safety of the vehicle. The \emph{efficiency} is defined as the average velocity of the ego vehicle.

The proposed method and the STV method \cite{zhang2021unified} proposed by Zhang \emph{et al.} are used for testing. We implement the STV method and apply it for real-time trajectory planning. At the same time, the ego vehicle's corresponding agent's states from the dataset are also recorded as the human driver. The testing results are shown in Tab.~\ref{tab:base_compare}. In the test, the average time consumption of the proposed method's one single planning episode is 58.3 ms, which fully meets the real-time requirements.

\begin{table}[!t]
    \centering
    \renewcommand{\arraystretch}{1.5}
    \fontsize{8}{8}\selectfont
    \begin{threeparttable}
        \caption{Comparison of different methods in the open-loop test.} 
        \label{tab:base_compare}
        \setlength{\tabcolsep}{2.5pt}
        \begin{tabular}{ccccccccc}
            \toprule
            \multirow{2}{*}{Method} & \multicolumn{4}{c}{Lane Keeping} & \multicolumn{4}{c}{Lane Change} \cr
            \cmidrule(lr){2-5} \cmidrule(lr){6-9}
            & Succ. $\uparrow$ & Fail $\downarrow$ & Risk $\downarrow$ & Effi. $\uparrow$ & Succ. $\uparrow$ & Fail $\downarrow$ & Risk $\downarrow$ & Effi. $\uparrow$ \cr
            \midrule
            STV & 56\% & 44\% & 45.9\% & 11.76 & 0\% & 83\% & 50.9\% & 15.30 \cr
            Human & - & - & 25.8\% & 12.41 & - & - & 52.4\% & 16.29 \cr
            Ours & \textbf{91\%} & \textbf{9\%} & \textbf{10.2\%} & \textbf{12.74} & \textbf{45\%} & \textbf{24\%} & \textbf{23.7\%} & \textbf{17.11} \cr
            \bottomrule
        \end{tabular}
    \end{threeparttable}
\end{table}

\begin{figure}[!t]
    \centering
    \includegraphics[width=\linewidth, trim=50 160 50 160, clip]{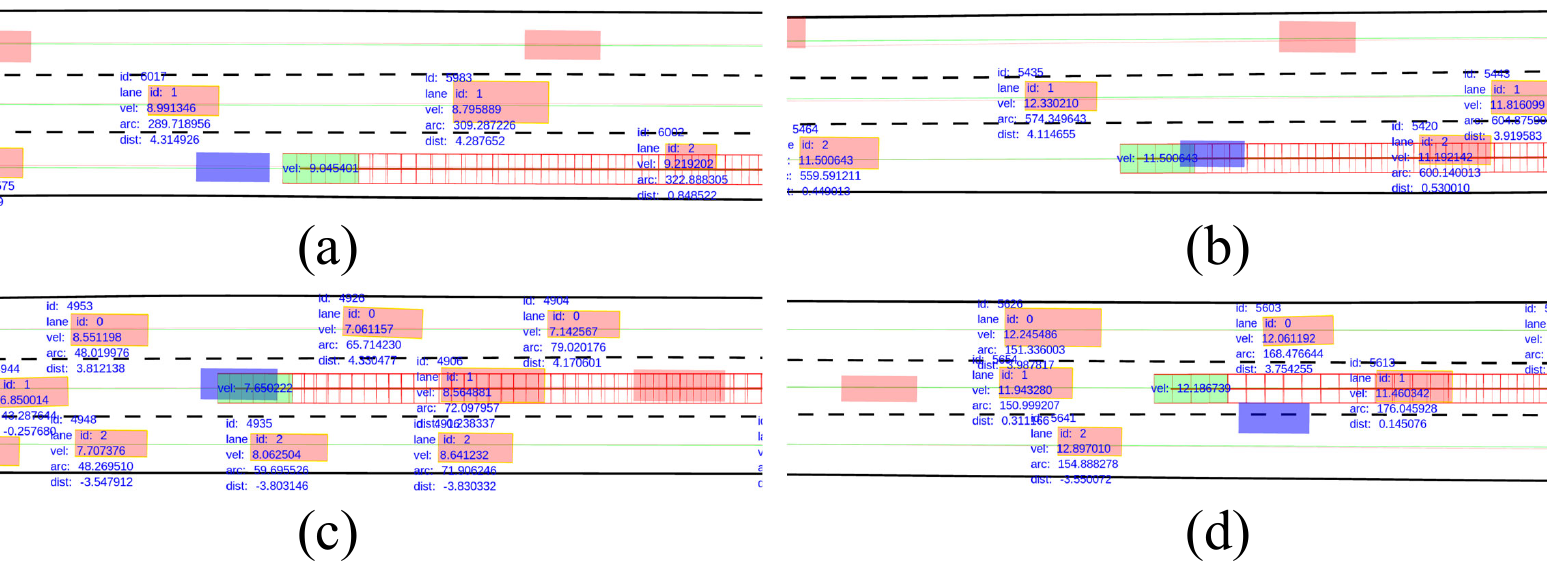}
    \caption{This figure shows four screenshots in the open-loop tests. The red boxes indicate the agents. The blue box indicates the ego vehicle operated by the human driver. The green box indicates the ego vehicle using the proposed method. Fig.~(a), (b), and (c) are the lane keeping and Fig.~(d) is the lane changing.}
    \label{fig:open_loop_test}
    \vspace{-0.2in}
\end{figure}

It can be seen that according to each measurement standard, the proposed methods outperform the STV methods in all the metrics. The reasons for the STV method's poor performance in the test mainly include three points. The first reason is that the traffic flow in the CitySim dataset is too dense and the ego vehicle cannot maintain a sufficient distance from other agents, so the STV method is difficult to generate the required voxels, causing planning failure. The second reason is that the STV method does not consider the lateral kinodynamic constraints in the voxels generation, so it cannot perform real-time trajectory planning in lane changing. The third reason is that the STV method's objective function is too naive and constraints are too strict, which makes it difficult to achieve better results in trajectory optimization. The proposed method solves the above problems and makes the effect of the planning method greatly improved.

Compared with the human driver, the proposed method also achieves lower risk and higher efficiency. Fig.~\ref{fig:open_loop_test} shows four examples of the comparison between the proposed method and the human driver. Fig.~\ref{fig:open_loop_test} (a), (b), and (c) are the screenshots of the lane-keeping tests. Fig.~\ref{fig:open_loop_test} (a) shows a sparse environment, where the proposed method makes the ego vehicle travel faster than the human driver to achieve higher efficiency. Fig.~\ref{fig:open_loop_test} (b) shows a little bit dense environment, where the proposed method prefers to keep a long distance to the front agent to achieve lower risk. Fig.~\ref{fig:open_loop_test} (c) shows a very dense environment, where the proposed method almost performs the same as the human driver. These three cases can illustrate how the proposed method can achieve lower risk and higher efficiency than the human driver. Fig.~\ref{fig:open_loop_test} (d) shows a case that a human driver conducts a dangerous lane change in the dense traffic flow, where the proposed method fails to conduct the same behavior. In the lane-changing test scenario, the reason for the low success rate of the proposed method is that the traffic flow in the CitySim dataset is dense, and lane-changing often requires an aggressive strategy. However, the proposed method uses a more cautious strategy due to the lack of accurate prediction of the other agents' behaviors. As a result, in some cases, the proposed method is more likely to let the vehicle maintain the lane.

\subsection{Close-Loop Tests}

In the closed-loop test, the gym simulation \cite{highway-env} is applied. A four-lane highway scene is constructed through gym simulation, and other agents are randomly placed in it. The ego vehicle's maximum longitudinal velocity is set to 20 m/s, and that of other agents is set to 15 m/s. The ego vehicle runs continuously for 8 minutes using the proposed method, and the vehicle's longitudinal and lateral velocities during the process are recorded.

\begin{figure}[!t]
    \centering
    \includegraphics[width=\linewidth]{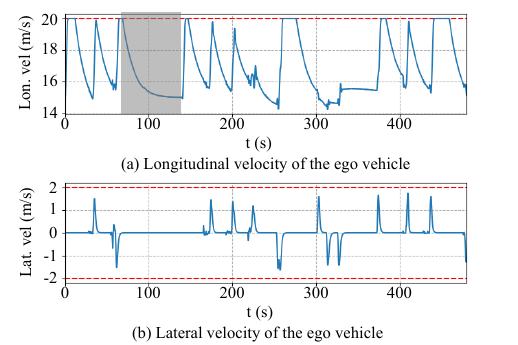}
    \caption{This figure shows the longitudinal and lateral velocities of the ego vehicle during the 8 minutes close-loop tests. The red lines indicate the longitudinal and lateral velocity limitations. The shadowed region indicates an example of vehicle following.}
    \label{fig:close_loop_test_all}
    \vspace{-0.2in}
\end{figure}

The vehicle finally passes the test successfully. Within 8 minutes, the vehicle safely interacted with other agents on the scene and maintained a high velocity. During the period, the vehicle successfully completed several times of smooth following and changing lanes to overtake. The vehicle's longitudinal and lateral velocity during the whole test process is shown in Fig.~\ref{fig:close_loop_test_all}.

From Fig.~\ref{fig:close_loop_test_all}, it can be seen that the ego vehicle's longitudinal and lateral velocities meet the constraints. During the test, the ego vehicle changes lanes and overtakes 12 times, and its average longitudinal velocity reaches 16.86 m/s, which is higher than the other agents' maximum velocity. It shows that when using the proposed method, the ego vehicle will not be hindered by the other low-velocity agents, and can travel with higher efficiency. Especially, the gray area in Fig.~\ref{fig:close_loop_test_all} is the ego vehicle's longitudinal velocity profile during a follow-up process lasting about 1 minute. It can be seen that when following a front agent using our planning method, the ego vehicle can decelerate smoothly until it reaches a steady state at the same velocity as the front agent. In contrast, the STV method is difficult to achieve stable vehicle following, and there will be frequent acceleration and deceleration during the vehicle following process. The reason is that the change of the objective function and the constraints on the position and velocity of the front agent are continuous in our defined optimization problem but discontinuous in the STV method.

\begin{figure}[!t]
    \centering
    \includegraphics[width=\linewidth, trim=50 0 50 0, clip]{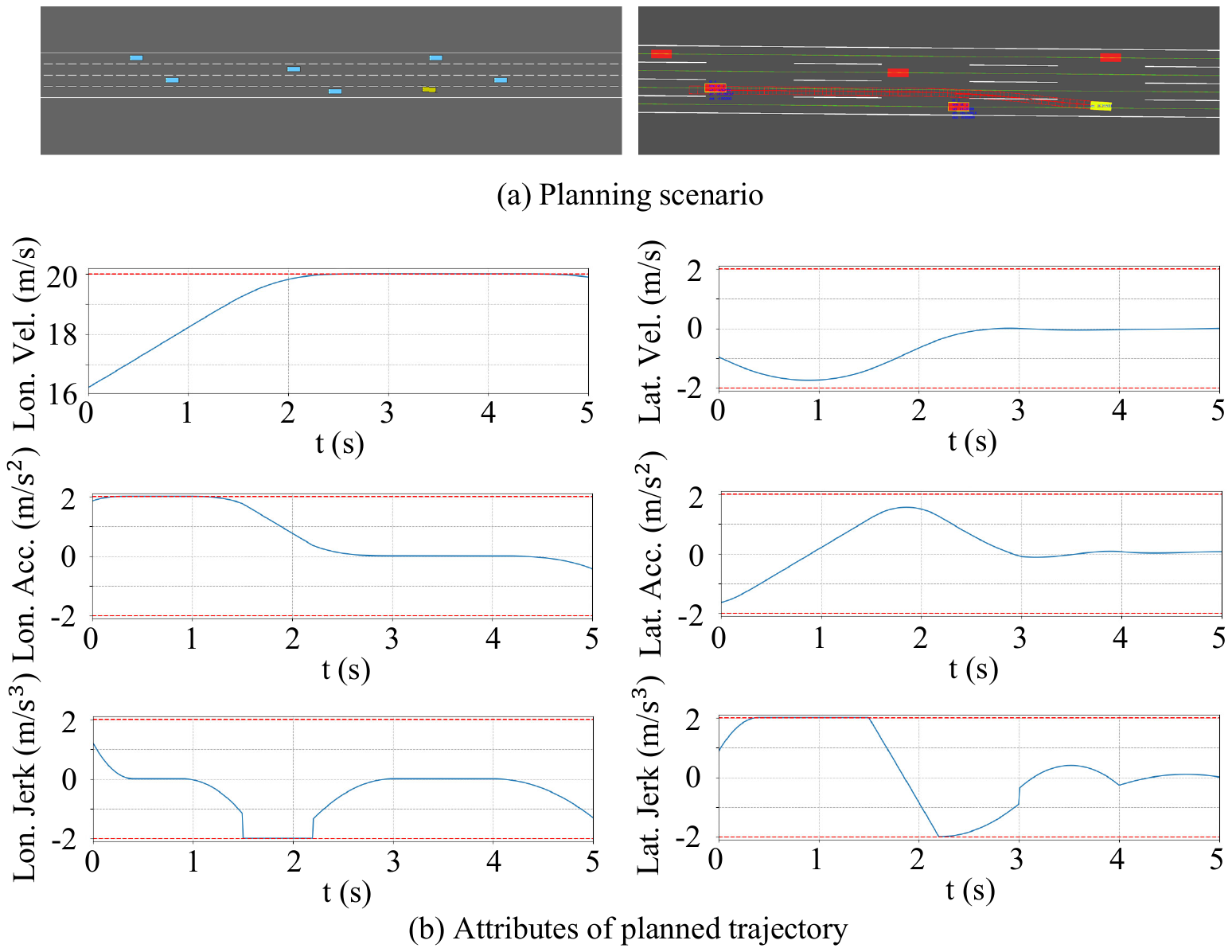}
    \caption{This figure shows an example of the planned trajectory during lane changing. (a) The left picture shows the visualization of the simulator and the right picture shows the visualization of the planning program. The yellow box indicates the ego vehicle, the blue and red boxes indicate the other agent. The red curve indicates the planned trajectory. (b) The blue profiles are the attributes of the planned trajectory. The red lines indicate the limitations.}
    \label{fig:close_loop_test_single}
    \vspace{-0.2in}
\end{figure}

In order to further demonstrate the proposed method, one single planning episode's result is in Fig.~\ref{fig:close_loop_test_single}. Fig.~\ref{fig:close_loop_test_single} (a) shows the scenario of this planning episode. Due to the obstruction of a low-speed agent in front and no other agents on the right lane, the proposed method plans the right lane-changing trajectory for overtaking. Fig.~\ref{fig:close_loop_test_single} (b) shows the properties of the planned trajectory. It can be seen that the planned trajectory's longitudinal and lateral velocities, accelerations, and jerks are continuous, which proves that the planned trajectory by our method satisfies the $\mathrm{C}^2$ continuity. In addition, the planned trajectory's longitudinal and lateral velocities, accelerations, and jerks all meet the constraints, which proves that the trajectory planned by our method can meet the needs of comfort.

\subsection{Ablation Study}

The ablation study uses the same process as the open-loop tests. In the ablation study, we fix the number of voxels in the voxel sequence as the first comparison method. The voxels' sizes in the time domain are fixed, that is, $\Delta T_{i+1} = \Delta T_i, \forall i$, as the second comparison method. Only the jerks are considered in the objective function, that is, $w_j=0, j=1,2,3,4$, as the third comparison method. Only the jerks and end states are considered in the objective function, that is, $w_j=0, j=3,4$, as the fourth comparison method. These four methods are used to test separately, and compared with the results obtained by the original method, as shown in Tab.~\ref{tab:ablation_study}.

It can be seen from the table that when the number of voxels is fixed, the method is worse than the original method under all metrics, which can prove that the introduction of the changeable voxel number is effective. Then, when the size of voxels in the time domain is fixed, other metrics except risk are worse than the original method. The reason for this result is that the size of the voxel in the time domain is positively correlated with the vehicle's response time. Fixing the size of voxels in the time domain will make the vehicle's response time as close as possible to this size, and when the vehicle's response time cannot approach this size, it will lead to planning failure. This characteristic causes the vehicle to either be in safety or planning fail. That's why using fixed-size voxels leads to a high failure rate and low risk. The original method solves this problem using multi-resolution voxels. Besides, the results of the 3rd and 4th comparing methods prove that the original method can improve the planner by introducing new items in the objective function.

\begin{table}[t]
    \centering
    \renewcommand{\arraystretch}{1.5}
    \fontsize{8}{8}\selectfont
    \begin{threeparttable}
        \caption{Comparison of different methods in the ablation study.} 
        \label{tab:ablation_study}
        \setlength{\tabcolsep}{2.0pt}
        \begin{tabular}{ccccccccc}
            \toprule
            \multirow{2}{*}{Method} & \multicolumn{4}{c}{Lane Keeping} & \multicolumn{4}{c}{Lane Change} \cr
            \cmidrule(lr){2-5} \cmidrule(lr){6-9}
            & Succ. $\uparrow$ & Fail $\downarrow$ & Risk $\downarrow$ & Effi. $\uparrow$ & Succ. $\uparrow$ & Fail $\downarrow$ & Risk $\downarrow$ & Effi. $\uparrow$ \cr
            \midrule
            Comp. I & 81\% & 19\% & 10.5\% & 12.53 & 3\% & 60\% & 26.3\% & 15.42 \cr
            Comp. II & 82\% & 18\% & \textbf{4.9\%} & 12.41 & 20\% & 62\% & \textbf{19.2\%} & 16.26 \cr
            Comp. III & 59\% & 41\% & 18.9\% & 11.65 & 16\% & 58\% & 28.1\% & 15.31 \cr
            Comp. IV & \textbf{91\%} & \textbf{9\%} & 10.2\% & 12.63 & 43\% & 26\% & 23.5\% & 16.10 \cr
            Origin & \textbf{91\%} & \textbf{9\%} & 10.2\% & \textbf{12.74} & \textbf{45\%} & \textbf{24\%} & 23\% & \textbf{17.11} \cr
            \bottomrule
        \end{tabular}
    \end{threeparttable}
    \vspace{-0.2in}
\end{table}

%% file: pages/conclusion.tex
\section{CONCLUSIONS}

In this paper, a variable-length, multi-resolution voxel sequence is used to represent highway scenes, and a new optimization problem is defined to describe the trajectory planning problem. Combining the above two innovations, this paper implements an environment-adaptive trajectory planning method, which can enable autonomous vehicles to achieve efficient, safe and comfortable traveling in highway scenarios. Experiments prove that the proposed method can perform better than the current planning methods. In future work, on the one hand, we will expand the applicable scenarios of this method so that it can be applied to urban environments. On the other hand, we plan to combine this method with the prediction method\cite{sun2022pseudo, sun2019interactive}, introduce the motion uncertainty of other agents\cite{sun2021diverse}, and further improve the effect of the method.

%% file: root.bbl
\begin{thebibliography}{10}
\providecommand{\url}[1]{#1}
\csname url@rmstyle\endcsname
\providecommand{\newblock}{\relax}
\providecommand{\bibinfo}[2]{#2}
\providecommand\BIBentrySTDinterwordspacing{\spaceskip=0pt\relax}
\providecommand\BIBentryALTinterwordstretchfactor{4}
\providecommand\BIBentryALTinterwordspacing{\spaceskip=\fontdimen2\font plus
\BIBentryALTinterwordstretchfactor\fontdimen3\font minus
  \fontdimen4\font\relax}
\providecommand\BIBforeignlanguage[2]{{%
\expandafter\ifx\csname l@#1\endcsname\relax
\typeout{** WARNING: IEEEtran.bst: No hyphenation pattern has been}%
\typeout{** loaded for the language `#1'. Using the pattern for}%
\typeout{** the default language instead.}%
\else
\language=\csname l@#1\endcsname
\fi
#2}}

\bibitem{claussmann2019review}
L.~Claussmann, M.~Revilloud, D.~Gruyer, and S.~Glaser, ``A review of motion
  planning for highway autonomous driving,'' \emph{IEEE Transactions on
  Intelligent Transportation Systems}, vol.~21, no.~5, pp. 1826--1848, 2019.

\bibitem{ding2019safe}
W.~Ding, L.~Zhang, J.~Chen, and S.~Shen, ``Safe trajectory generation for
  complex urban environments using spatio-temporal semantic corridor,''
  \emph{IEEE Robotics and Automation Letters}, vol.~4, no.~3, pp. 2997--3004,
  2019.

\bibitem{zhang2021unified}
T.~Zhang, W.~Song, M.~Fu, Y.~Yang, X.~Tian, and M.~Wang, ``A unified framework
  integrating decision making and trajectory planning based on spatio-temporal
  voxels for highway autonomous driving,'' \emph{IEEE Transactions on
  Intelligent Transportation Systems}, vol.~23, no.~8, pp. 10\,365--10\,379,
  2021.

\bibitem{https://doi.org/10.48550/arxiv.2208.11036}
\BIBentryALTinterwordspacing
O.~Zheng, M.~Abdel-Aty, L.~Yue, A.~Abdelraouf, Z.~Wang, and N.~Mahmoud,
  ``Citysim: A drone-based vehicle trajectory dataset for safety oriented
  research and digital twins,'' 2022. [Online]. Available:
  \url{https://arxiv.org/abs/2208.11036}
\BIBentrySTDinterwordspacing

\bibitem{highway-env}
E.~Leurent, ``An environment for autonomous driving decision-making,''
  \url{https://github.com/eleurent/highway-env}, 2018.

\bibitem{gonzalez2015review}
D.~Gonz{\'a}lez, J.~P{\'e}rez, V.~Milan{\'e}s, and F.~Nashashibi, ``A review of
  motion planning techniques for automated vehicles,'' \emph{IEEE Transactions
  on intelligent transportation systems}, vol.~17, no.~4, pp. 1135--1145, 2015.

\bibitem{ziegler2014trajectory}
J.~Ziegler, P.~Bender, T.~Dang, and C.~Stiller, ``Trajectory planning for
  bertha—a local, continuous method,'' in \emph{2014 IEEE intelligent
  vehicles symposium proceedings}.\hskip 1em plus 0.5em minus 0.4em\relax IEEE,
  2014, pp. 450--457.

\bibitem{werling2010optimal}
M.~Werling, J.~Ziegler, S.~Kammel, and S.~Thrun, ``Optimal trajectory
  generation for dynamic street scenarios in a frenet frame,'' in \emph{2010
  IEEE International Conference on Robotics and Automation}.\hskip 1em plus
  0.5em minus 0.4em\relax IEEE, 2010, pp. 987--993.

\bibitem{chu2012local}
K.~Chu, M.~Lee, and M.~Sunwoo, ``Local path planning for off-road autonomous
  driving with avoidance of static obstacles,'' \emph{IEEE transactions on
  intelligent transportation systems}, vol.~13, no.~4, pp. 1599--1616, 2012.

\bibitem{li2017development}
X.~Li, Z.~Sun, D.~Cao, D.~Liu, and H.~He, ``Development of a new integrated
  local trajectory planning and tracking control framework for autonomous
  ground vehicles,'' \emph{Mechanical Systems and Signal Processing}, vol.~87,
  pp. 118--137, 2017.

\bibitem{zhang2021trajectory}
J.~Zhang, Z.~Jian, J.~Fu, Z.~Nan, J.~Xin, and N.~Zheng, ``Trajectory planning
  with comfort and safety in dynamic traffic scenarios for autonomous
  driving,'' in \emph{2021 IEEE Intelligent Vehicles Symposium Workshops (IV
  Workshops)}.\hskip 1em plus 0.5em minus 0.4em\relax IEEE, 2021, pp. 342--349.

\bibitem{xu2012real}
W.~Xu, J.~Wei, J.~M. Dolan, H.~Zhao, and H.~Zha, ``A real-time motion planner
  with trajectory optimization for autonomous vehicles,'' in \emph{2012 IEEE
  International Conference on Robotics and Automation}.\hskip 1em plus 0.5em
  minus 0.4em\relax IEEE, 2012, pp. 2061--2067.

\bibitem{villagra2012smooth}
J.~Villagra, V.~Milan{\'e}s, J.~P{\'e}rez, and J.~Godoy, ``Smooth path and
  speed planning for an automated public transport vehicle,'' \emph{Robotics
  and Autonomous Systems}, vol.~60, no.~2, pp. 252--265, 2012.

\bibitem{fan2018baidu}
H.~Fan, F.~Zhu, C.~Liu, L.~Zhang, L.~Zhuang, D.~Li, W.~Zhu, J.~Hu, H.~Li, and
  Q.~Kong, ``Baidu apollo em motion planner,'' \emph{arXiv preprint
  arXiv:1807.08048}, 2018.

\bibitem{lim2019hybrid}
W.~Lim, S.~Lee, M.~Sunwoo, and K.~Jo, ``Hybrid trajectory planning for
  autonomous driving in on-road dynamic scenarios,'' \emph{IEEE Transactions on
  Intelligent Transportation Systems}, vol.~22, no.~1, pp. 341--355, 2019.

\bibitem{ziegler2009spatiotemporal}
J.~Ziegler and C.~Stiller, ``Spatiotemporal state lattices for fast trajectory
  planning in dynamic on-road driving scenarios,'' in \emph{2009 IEEE/RSJ
  International Conference on Intelligent Robots and Systems}.\hskip 1em plus
  0.5em minus 0.4em\relax IEEE, 2009, pp. 1879--1884.

\bibitem{ding2021epsilon}
W.~Ding, L.~Zhang, J.~Chen, and S.~Shen, ``Epsilon: An efficient planning
  system for automated vehicles in highly interactive environments,''
  \emph{IEEE Transactions on Robotics}, vol.~38, no.~2, pp. 1118--1138, 2021.

\bibitem{farin2014curves}
G.~Farin, \emph{Curves and surfaces for computer-aided geometric design: a
  practical guide}.\hskip 1em plus 0.5em minus 0.4em\relax Elsevier, 2014.

\bibitem{gertz2003object}
E.~M. Gertz and S.~J. Wright, ``Object-oriented software for quadratic
  programming,'' \emph{ACM Transactions on Mathematical Software (TOMS)},
  vol.~29, no.~1, pp. 58--81, 2003.

\bibitem{quigley2009ros}
M.~Quigley, K.~Conley, B.~Gerkey, J.~Faust, T.~Foote, J.~Leibs, R.~Wheeler,
  A.~Y. Ng, \emph{et~al.}, ``Ros: an open-source robot operating system,'' in
  \emph{ICRA workshop on open source software}, vol.~3, no. 3.2.\hskip 1em plus
  0.5em minus 0.4em\relax Kobe, Japan, 2009, p.~5.

\bibitem{sun2021move}
M.~Sun, F.~Baldini, P.~Trautman, and T.~Murphey, ``Move beyond trajectories:
  Distribution space coupling for crowd navigation,'' in \emph{Robotics:
  Science and Systems}.\hskip 1em plus 0.5em minus 0.4em\relax RSS Foundation,
  2021, pp. 1--12.

\bibitem{sun2022pseudo}
L.~Sun, C.~Tang, Y.~Niu, E.~Sachdeva, C.~Choi, T.~Misu, M.~Tomizuka, and
  W.~Zhan, ``Domain knowledge driven pseudo labels for interpretable
  goal-conditioned interactive trajectory prediction,'' in \emph{2022 IEEE/RSJ
  International Conference on Intelligent Robots and Systems (IROS)}, 2022, pp.
  13\,034--13\,041.

\bibitem{sun2019interactive}
L.~Sun, W.~Zhan, D.~Wang, and M.~Tomizuka, ``Interactive prediction for
  multiple, heterogeneous traffic participants with multi-agent hybrid dynamic
  bayesian network,'' in \emph{2019 IEEE Intelligent Transportation Systems
  Conference (ITSC)}, 2019, pp. 1025--1031.

\bibitem{sun2021diverse}
Z.-H. Yin, L.~Sun, L.~Sun, M.~Tomizuka, and W.~Zhan, ``Diverse critical
  interaction generation for planning and planner evaluation,'' in \emph{2021
  IEEE/RSJ International Conference on Intelligent Robots and Systems (IROS)},
  2021, pp. 7036--7043.

\end{thebibliography}
